\documentclass[conference]{IEEEtran}
\IEEEoverridecommandlockouts
\usepackage{cite}
\usepackage{amsmath,amssymb,amsfonts}
\usepackage{algorithmic}
\usepackage{graphicx}
\usepackage{textcomp}
\usepackage{xcolor}
\usepackage[inkscapelatex]{svg}
\usepackage{float}
\usepackage{subfigure}
\usepackage{soul}
\usepackage{color,xcolor}
\def\BibTeX{{\rm B\kern-.05em{\sc i\kern-.025em b}\kern-.08em
    T\kern-.1667em\lower.7ex\hbox{E}\kern-.125emX}}

\makeatletter
\newcommand{\linebreakand}{%
  \end{@IEEEauthorhalign}
  \hfill\mbox{}\par
  \mbox{}\hfill\begin{@IEEEauthorhalign}
}
\makeatother

\begin{document}

\title{Group Equivariant BEV for 3D Object Detection\
\thanks{This research is supported by National Natural Science Foundation of China (No.42130112) and KartoBit Research Network(No.KRN2201CA). Partially supported by ‘Fujian Science \& Technology Innovation Laboratory for Optoelectronic Information of China’ (No.2021ZZ120).
}
 \thanks{*Corresponding Author}
}

\author{\IEEEauthorblockN{1\textsuperscript{st} Hongwei Liu}
\IEEEauthorblockA{\textit{Fujian Institute of Research on} \\
\IEEEauthorblockA{\textit{the Structure of Matter}}
\textit{Chinese Academy of Sciences}\\
Fuzhou, China \\
hwhongwei.liu@foxmail.com}
\and
\IEEEauthorblockN{2\textsuperscript{nd} Jian Yang$^{\ast}$}
\IEEEauthorblockA{\textit{School of Geospatial Information} \\
\textit{Information Engineering University}\\
Zhengzhou, China \\
jian.yang@tum.de}
\and
\IEEEauthorblockN{3\textsuperscript{rd} Jianfeng Zhang}
\IEEEauthorblockA{\textit{Fujian Institute of Research on} \\
\IEEEauthorblockA{\textit{the Structure of Matter}}
\textit{Chinese Academy of Sciences}\\
Fuzhou, China \\
zhangjf@fjirsm.ac.cn}
\linebreakand
\IEEEauthorblockN{4\textsuperscript{th} Dongheng Shao}
\IEEEauthorblockA{\textit{Fujian Institute of Research on} \\
\IEEEauthorblockA{\textit{the Structure of Matter}}
\textit{Chinese Academy of Sciences}\\
Fuzhou, China \\
shaodongheng@fjirsm.ac.cn}
\and
\IEEEauthorblockN{5\textsuperscript{th} Jielong Guo}
\IEEEauthorblockA{\textit{Fujian Institute of Research on} \\
\IEEEauthorblockA{\textit{the Structure of Matter}}
\textit{Chinese Academy of Sciences}\\
Fuzhou, China \\
gjl@fjirsm.ac.cn}
\and
\IEEEauthorblockN{6\textsuperscript{th} Shaobo Li}
\IEEEauthorblockA{\textit{School of Electrical Engineer} \\
\IEEEauthorblockA{\textit{and Automation}} 
\textit{Xiamen University of Technology}\\
Xiamen, China \\
lishaobogo@gmail.com}
\linebreakand
\IEEEauthorblockN{7\textsuperscript{th} Xuan Tang}
\IEEEauthorblockA{\textit{Fujian Institute of Research on} \\
\IEEEauthorblockA{\textit{the Structure of Matter}}
\textit{Chinese Academy of Sciences}\\
Fuzhou, China \\
xtang@cee.ecnu.edu.cn}
\and
\IEEEauthorblockN{8\textsuperscript{th} Xian Wei}
\IEEEauthorblockA{\textit{Fujian Institute of Research on} \\
\IEEEauthorblockA{\textit{the Structure of Matter}}
\textit{Chinese Academy of Sciences}\\
Fuzhou, China \\
xian.wei@tum.de}
}

% \author{
%     \IEEEauthorblockN{Hongwei Liu$^{1}$, Jian Yang$^{2{\ast}}$, Jianfeng Zhang$^{1}$, Dongheng Shao$^{1}$, \\Jielong Guo$^{1}$, Shaobo Li$^{3}$, Xuan Tang$^{1}$, Xian Wei$^{1}$}
%     \IEEEauthorblockA{$^1$ Fujian Institute of Research on the Structure of Matter, Chinese Academy of Sciences, Fuzhou, China}
%     \IEEEauthorblockA{$^2$ School of Geospatial Information, Information Engineering University, Zhengzhou, China}
%     \IEEEauthorblockA{$^3$ School of Electrical Engineer and Automation, Xiamen University of Technology, Xiamen, China}
%     \IEEEauthorblockA{\{hwhongwei.liu\}@foxmail.com, \{jian.yang, xian.wei\}@tum.de, \{zhangjf, shaodongheng, gjl\}@fjirsm.ac.cn}
%     \IEEEauthorblockA{\{lishaobogo\}@gmail.com, \{xtang\}@cee.ecnu.edu.cn}
% }

\maketitle

\begin{abstract}
Recently, 3D object detection has attracted significant attention and achieved continuous improvement in real road scenarios. The environmental information is collected from a single sensor or multi-sensor fusion to detect interested objects. However, most of the current 3D object detection approaches focus on developing advanced network architectures to improve the detection precision of the object rather than considering the dynamic driving scenes, where data collected from sensors equipped in the vehicle contain various perturbation features. As a result, existing work cannot still tackle the perturbation issue. In order to solve this problem, we propose a group equivariant bird’s eye view network (GeqBevNet) based on the group equivariant theory, which introduces the concept of group equivariant into the BEV fusion object detection network. The group equivariant network is embedded into the fused BEV feature map to facilitate the BEV-level rotational equivariant feature extraction, thus leading to lower average orientation error. In order to demonstrate the effectiveness of the GeqBevNet, the network is verified on the nuScenes validation dataset in which mAOE can be decreased to $0.325$. Experimental results demonstrate that GeqBevNet can extract more rotational equivariant features in the 3D object detection of the actual road scene and improve the performance of object orientation prediction.
\end{abstract}

% \begin{IEEEkeywords}
% group equivariant, BEV, 3D object detection, rotational equivariant feature
% \end{IEEEkeywords}

\section{Introduction}
%As a core part 
As one of the critical components of self-driving cars \cite{b1} and autonomous mobile robots \cite{b2}, perception systems have witnessed continuous development in recent years. In order to guarantee an application-ready performance of environmental perception, perception systems often employ various sensors to obtain environmental information. For example, LiDAR uses the optical time-of-flight(TOF) method to obtain point clouds with distance and geometric information through laser beams, which provide outlines and position information of its surrounding objects. However, several defects still limit LiDAR's further applications, such as the high cost, sparse point clouds of distant objects, and lack of semantics. Compared with LiDAR, cameras have been widely used in perception systems \cite{b6},\cite{b7} in real-world applications, with mature technology and low cost. It can provide rich semantic information, such as the color and texture of objects, and it can also recognize traffic lights and signboards in outdoor scenes. However, many uncertainties remain under extreme driving conditions, e.g., rainy, snowy, and too-bright weather conditions. It is difficult to extract sufficient context information \cite{b8} from dim or too-bright images. Therefore, multi-sensor fusion techniques are favored, which can safely and efficiently perform environment perception tasks.

Extensive research has recently investigated object detection networks based on multi-sensor fusion. The multi-modal information provided by multi-sensors can effectively utilize the advantages of each sensor to provide safe and reliable perception information \cite{b10},\cite{b11},\cite{b12},\cite{b13}. However, most of these sensor fusion research chose one sensor as the dominant data source while using the other sensor for supplementary information. For example, the work in \cite{b10} used camera inputs to acquire the image features from the front view while projecting the point cloud into the front view for sensor fusion. This approach makes little use of the geometric features of the point cloud. The other work \cite{b11} explored the geometric features in the LiDAR point clouds, then attached the semantic features of the images to the point clouds to perform object detection on the point clouds, which discards the semantic density in the images. Unifying sensory inputs from cameras and LiDAR in the same form for data fusion have become an important research topic of multi-sensor fusion. Recently, two important research work \cite{b14},\cite{b15} have contributed to such effort and employ a bird's eye view (BEV) for feature fusion. The BEV framework has enabled a unified top-down representation of sensory inputs from different modalities and alleviates the issue of object occlusion.

The BEV-based fusion methods overcome the shortcomings of previous fusion methods that cannot leverage both semantic and geometric information, thus resulting in a better performance in object detection. However, they have not considered the sensory perturbation of moving objects and the varying road scenes. Detecting moving objects with various perturbation characteristics is a non-trivial task that has not been well studied, especially in developing advanced neural network architectures robust to varying perturbation. Data augmentation is an alternative method to improve the detection performance for moving objects, but it relies on sufficient training data and requires large computational overhead. In contrast to data augmentation, another work\cite{b3} used group equivariant methods to solve the data redundancy problem of data augmentation, in addition to alleviating the performance degradation caused by vehicle rotation, which is a highly worthwhile idea.

To address the above issues, we propose a group equivariant BEV object detection network named group equivariant bird’s eye view network($\mathbf{GeqBevNet}$). A group equivariant network is embedded into a 3D detection framework to extract rotational equivariant features. Specifically, the camera BEV generation network and the LiDAR BEV generation network generate two BEV-based sensory input representations separately and fuse these two BEVs. Furthermore, the group equivariant network is embedded after the fused BEV and extracts the rotational equivariant features using group operations. We achieve excellent experimental results on the challenging benchmark dataset nuScenes \cite{b17} for autonomous driving. On the NuScenes validation set, GeqBevNet achieves $62.2\%$ on mean Average Precision (mAP) and $0.325$ on mean Average Orientation Error (mAOE). Compared with other advanced networks, GeqBevNet has a significantly lower mAOE. It shows that our GeqBevNet performs orientation prediction better in scene rotation. The main contributions of this paper are as follows:

\begin{itemize}
    \item We construct an embeddable group equivariant network based on group equivariant theory to extract rotational equivariant features at BEV-level.
    \item To satisfied the requirement of embeddability, the group equivariant network can be freely and reasonably embedded into the object detection network.
    \item To demonstrate the effectiveness of the proposed GeqBevNet, we conducted extensive experiments on the nuScenes dataset, where we achieved better results compared to advanced object detection networks.
\end{itemize}

%第一幅图
\begin{figure*}[htbp]
\centerline{\includegraphics[width=18cm]{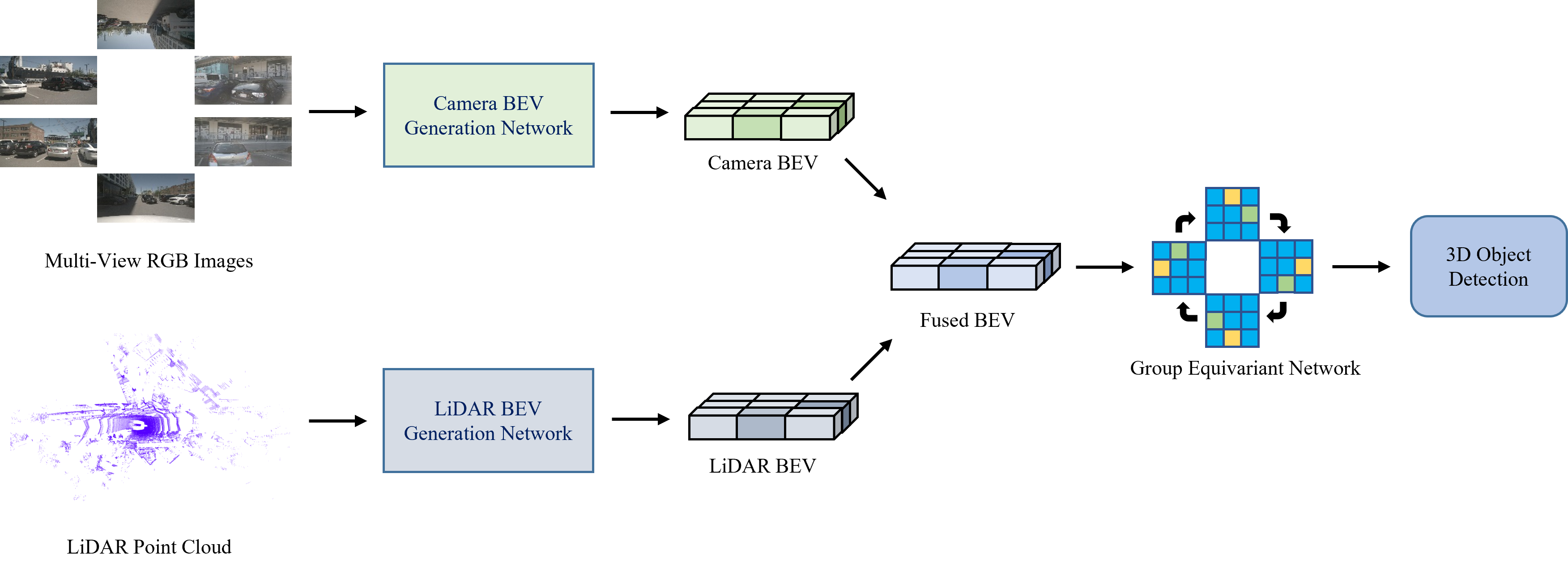}}
\caption{The overall framework of the proposed GeqBevNet network. GeqBevNet first generates BEV through the camera BEV generation network and LiDAR BEV generation networks and fuses them. After fusing the BEV, the group equivariant network is embedded to extract the rotational equivariant features.}
\label{fig1}
\end{figure*}

\section{RELATED WORK}

\subsection{3D Object Detection}
\subsubsection{Camera-based 3D Object Detection}

3D object detection directly on the front view is still a challenge. In the previous camera-based 3D object detection methods, one approach uses mature 2D object detection methods to extract region proposals on the front view, then return to 3D bounding boxes. Chen et al. \cite{b18} generated a set of 2D object region proposals from a single image and then used the proposed method to put object candidates into 3D. Kundu et al. \cite{b19} introduced the concept of CAD models to learn a low-dimensional shape space while extracting 2D region proposals. Another approach is to convert the front view representation, which can be done directly using existing LIDAR-based object detection methods. Wang et al. \cite{b20} performed depth estimation based on the front view to generate a pseudo-point cloud and used LIDAR-based target detection methods directly on the pseudo-point cloud. In contrast, Qian et al. \cite{b21} used a binocular camera to compute disparity for depth estimation, generating a pseudo-point cloud.  

Recently, BEV-based object detection method \cite{b22} has achieved SOTA on the public dataset nuScenes. Converting the front view to BEV facilitates object detection and dramatically alleviates the occlusion problem of object detection in the front view perspective. However, the quality of BEV generation seriously affects the target detection performance. Currently, there are two ways to convert the front view to BEV: one is to use the internal and external parameters of the camera explicitly, and the other is to use the internal and external parameters implicitly. Mallot et al. \cite{b23} proposed IPM, which directly uses internal and external parameters to convert BEV, provided all objects are on the plane. However, bumps can seriously affect the quality of BEV generation and cause distortion. Although LSS\cite{b24} avoids the disadvantage of IPM, it uses depth prediction to generate BEV, while depth prediction needs strong prior knowledge. In order to obtain accurate depth information, Li et al. \cite{b26} proposed BEVDepth, which introduces LiDAR's depth information to supervise the camera's depth prediction in the training stage, and the obtained depth information has higher accuracy. There are also methods for generating BEV that implicitly utilize information external to the camera and do not require depth prediction. Lu et al. \cite{b27} generated BEV using MLP strategy and used a fixed matrix to transform the front view, do not depend on the input image.
%lack an inference process, 
% and are only suitable for transforming a single image.
In contrast to MLP-based methods, BEV converting method proposed by Wang et al.\cite{b28} and Li et al.\cite{b29} is transformer-based and highly dependent on the input data.

\subsubsection{LiDAR-based 3D Object Detection}
The point cloud provided by LiDAR is a natural 3D space that provides accurate depth information. However, Due to the irregularity and disorder of point clouds, early work focused on directly processing raw point clouds. Qi et al.  \cite{b30} proposed PointNet as the foundational work to process raw point clouds directly. The follow-up PointNet++ \cite{b31} made up for the deficiency that PointNet does not extract local features and improves the model's generalization ability. Li et al. \cite{b48} defined an operation called $\mathcal{X} $-Conv, and regular convolution can process raw point clouds. However, the model's input data is cumbersome if the point cloud is performed directly. To reduce the amount of data, Zhou et al. \cite{b33} processed the raw point cloud into regular voxels, which can be directly used for 3D convolution. Unlike voxels that divide into three dimensions, PointPillars proposed by Lang et al. \cite{b34} split the point cloud into 2D point cloud columns, which are then converted into BEV. This approach dramatically improves the speed of point cloud processing.

\subsubsection{Camera-LiDAR fusion for 3D Object Detection}
Since the single modality information provided by a single sensor is insufficient for 3D target detection in complex scenes, and the information provided by the camera and LiDAR are complementary in most scenes, most multi-sensor fusion efforts are based on cameras and LiDAR. Previous multi-sensor fusion works can be divided into two categories. One is camera-to-LiDAR. This work\cite{b11} used the image's semantic features to enhance point clouds but lost semantic density. Another category is LiDAR-to-camera. This category of work project point clouds onto a 2D plane and then processes it using existing 2D object detection methods. Chen et al. \cite{b10} projected the point cloud into the front view, discarding the geometric features of the point cloud. In order to avoid the loss of semantic density or the loss of geometric features caused by fusion, the fusion of the camera and LiDAR in the same form becomes the focus of future fusion work. Recently, Liu et al.\cite{b15} proposed BEVFusion to fuse camera BEV with LiDAR BEV and achieved remarkable results in 3D object detection. However, Liang et al.\cite{b14} focused more on the robustness of multi-sensor fusion in their BEV fusion work to study the perception performance of occlusion or single-sensor failure.

%第二幅图
\begin{figure*}[htbp]
% \vspace{-1cm}   %调整图片与上文的垂直距离  
% \setlength{\abovecaptionskip}{1cm} %调整标题上方的距离   
% \setlength{\abovecaptionskip}{-2cm} %调整标题下方的距离 	
    \centerline{\includegraphics[width=18cm]{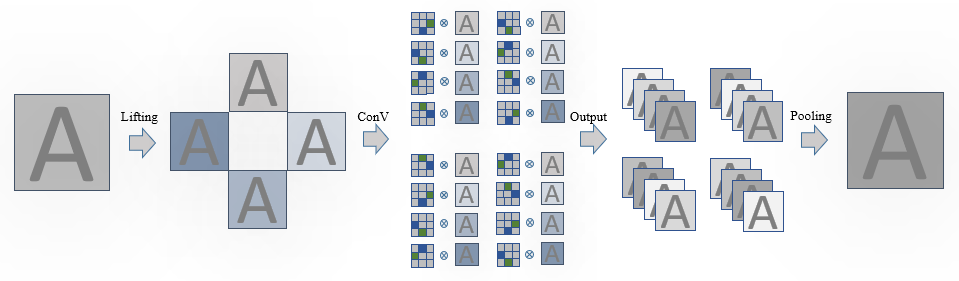}}
\caption{Process of extracting rotational equivariant features with group equivariant network.}
\label{fig2}
\end{figure*}

\subsection{Group Equivariant Network}
In traditional convolutional neural networks(CNN), which have translational equivariant in convolution, the object processing of an image can not be affected by the object's position in the image. However, it does not have the rotational equivariant. In previous work, rotational equivariant is usually obtained by simply rotating the training dataset images. Dillerman et al. \cite{b39} used rotational symmetry of images to predict the shape of early galaxies.

Group equivariant network is an equivariant network that uses group equivariant convolution to extract features with specific group properties. Cohen et al. \cite{b38} first proposed the group equivariant convolutional neural network, which lifted the traditional convolution to group with group properties. It was experimentally demonstrated that the group equivariant convolutional neural network has a higher degree of weight sharing and improves the network's performance without significantly increasing the number of network parameters. Since larger groups require higher computational cost, Cohen et al. \cite{b40} proposed a related theoretical framework to address this issue to separate group size from computational cost. Finzi et al. \cite{b41} proposed a general group equivariant convolution called LieConv, which allows a flexible alternative to a given group. Hutchinson et al. \cite{b42} introduced self-attention to build a LieSelfAttention layer with group equivariant properties.

Despite the  rapid progress in the study of group equivariant networks, it is still rarely used in 3D object detection or object recognition, and there are still significant challenges. Wang et al. \cite{b35} used equivariance as additional prior knowledge added to the network during continuous learning, which avoids an increase sharply in task increment caused by point cloud augmentation. Wu et al. \cite{b43} extract lightweight equivariant features on point clouds to improve real-time performance for 3D object detection. However, using rotational equivariant to solve the problems of 3d target detection on multi-sensor fusion target detection networks with rich data forms is still a challenge.

\section{Proposed Method}
\label{sec_3}

\subsection{Preliminary}
Mathematically, given that there are two domains, $X$ and $Y$, both belonging to the group $G$, then the definition of equivariance can be expressed as:
\begin{equation}
I(g \cdot x)=g \cdot I(x)\label{eq1}
\end{equation}
where $I$ is a function, $g$ belongs to $G$, $x$ belongs to X, and $g \cdot I(x)$ belongs to $Y$, and $X\to Y$ is called an equivariant mapping, and both X and Y are homomorphisms in the group $G$. For traditional CNN, translational equivariant means that the result of translation followed by convolution of the target in the image is consistent with the translation. As defined by Cohen et al. \cite{b38}, in traditional CNN, the definition of equivariance can be expressed as:
\begin{equation}
\left[L_{g} f\right](x)=f\left(g^{-1} x\right)\label{eq2}
\end{equation}
where $L{g}$  is a special group transform operator, and $x$ is expressed as an image feature map. 

% For traditional CNN, $Lg$ corresponds to the translational transformation. Equation \ref{eq2} means represents the object pixel translation and then mapping, which is equivalent to the object pixel mapping first and then the mapping feature translation, which satisfies the homomorphism defined by Cohen et al.\cite{b38}as follows:
% \begin{equation}
% L_{g}L_{h}=L_{gh}\label{eq3}
% \end{equation}
% where $L_{g}$ , $L_{h}$ and $L_{gh}$ are transformation operators of a specific discrete group.

Traditional CNN only has translational equivariant and cannot extract effective rotational orientation information. Previous work \cite{b38} proved that the traditional CNN does not have a rotational equivariant and proposed a related mathematical theory to enable CNN with a rotational equivariant. To achieve rotational equivariant for CNN, Cohen et al. \cite{b38} introduced the symmetry group $p4$ and lifted the input image to the symmetry group $p4$, which is equivalent to extending the function and convolution kernel defined on the plane $\mathbb{Z}^{2}  $ to the symmetry group $p4$. They expressed this process mathematically as follows:
\begin{equation}
[f \star  \psi](g)=\sum_{y \in \mathbb{Z}^{2}} \sum_{k} f_{k}(y) \psi_{k}\left(g^{-1} y\right)\label{eq3}
\end{equation}

The obtained image has specific properties of the symmetry group $p4$ and the convolution operation is performed on the symmetry group $p4$. This process can be defined by Cohen et al.\cite{b38} as follows:
\begin{equation}
[f \star  \psi](g)=\sum_{h \in G} \sum_{k} f_{k}(h) \psi_{k}\left(g^{-1} h\right)\label{eq4}
\end{equation}

With the above processing, the image features acquire rotational equivariance while having translational equivariance.

%第三幅图
\begin{figure*}[htbp]

\subfigure[Scene1]{
\begin{minipage}[t]{0.25\linewidth}
\centerline{\includegraphics[width=5cm]{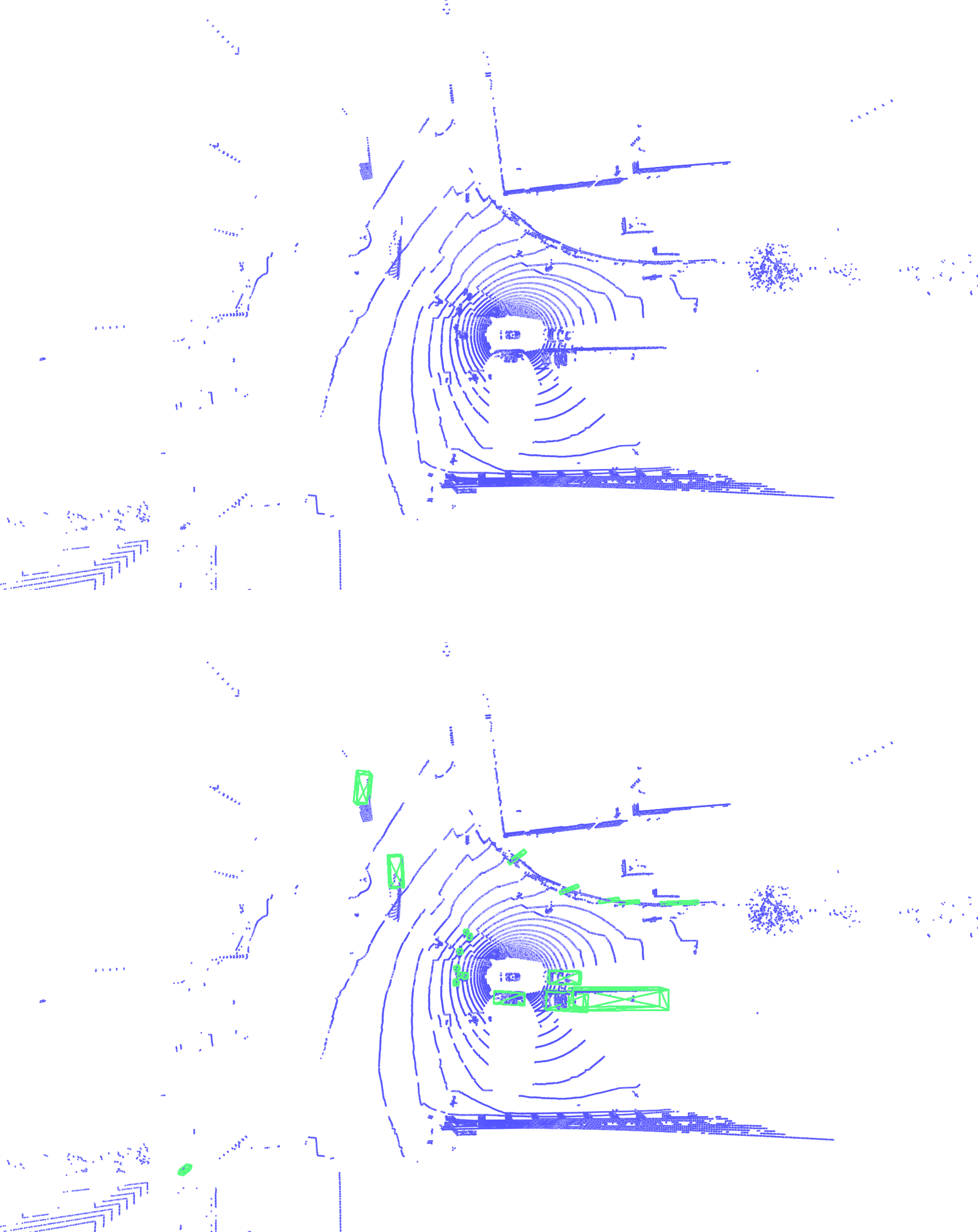}}
%\caption{fig1}
\end{minipage}%
}%
\subfigure[Scene2]{
\begin{minipage}[t]{0.25\linewidth}
\centerline{\includegraphics[width=5cm]{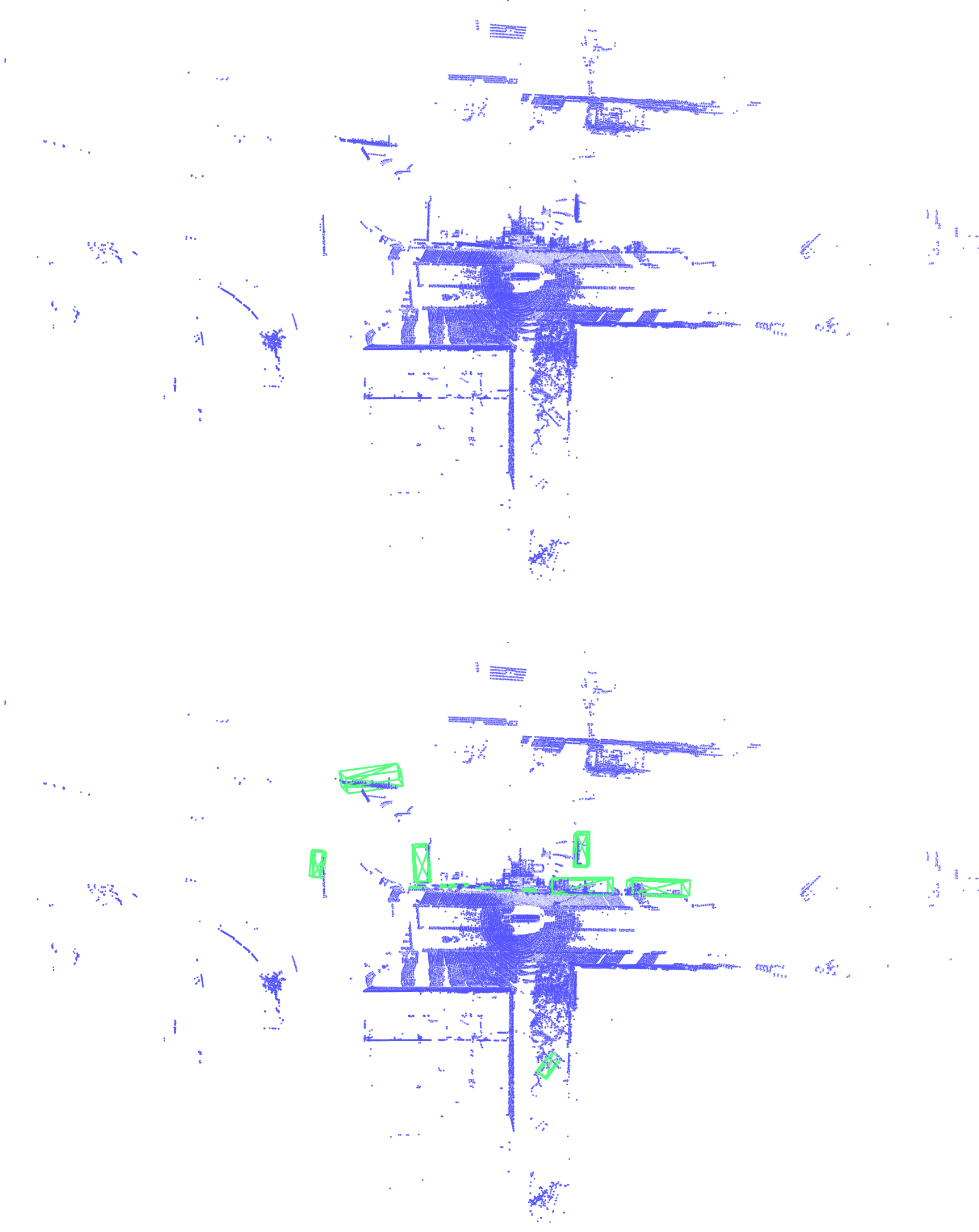}}
%\caption{fig2}
\end{minipage}%
}%
\subfigure[Scene3]{
\begin{minipage}[t]{0.25\linewidth}
\centerline{\includegraphics[width=5cm]{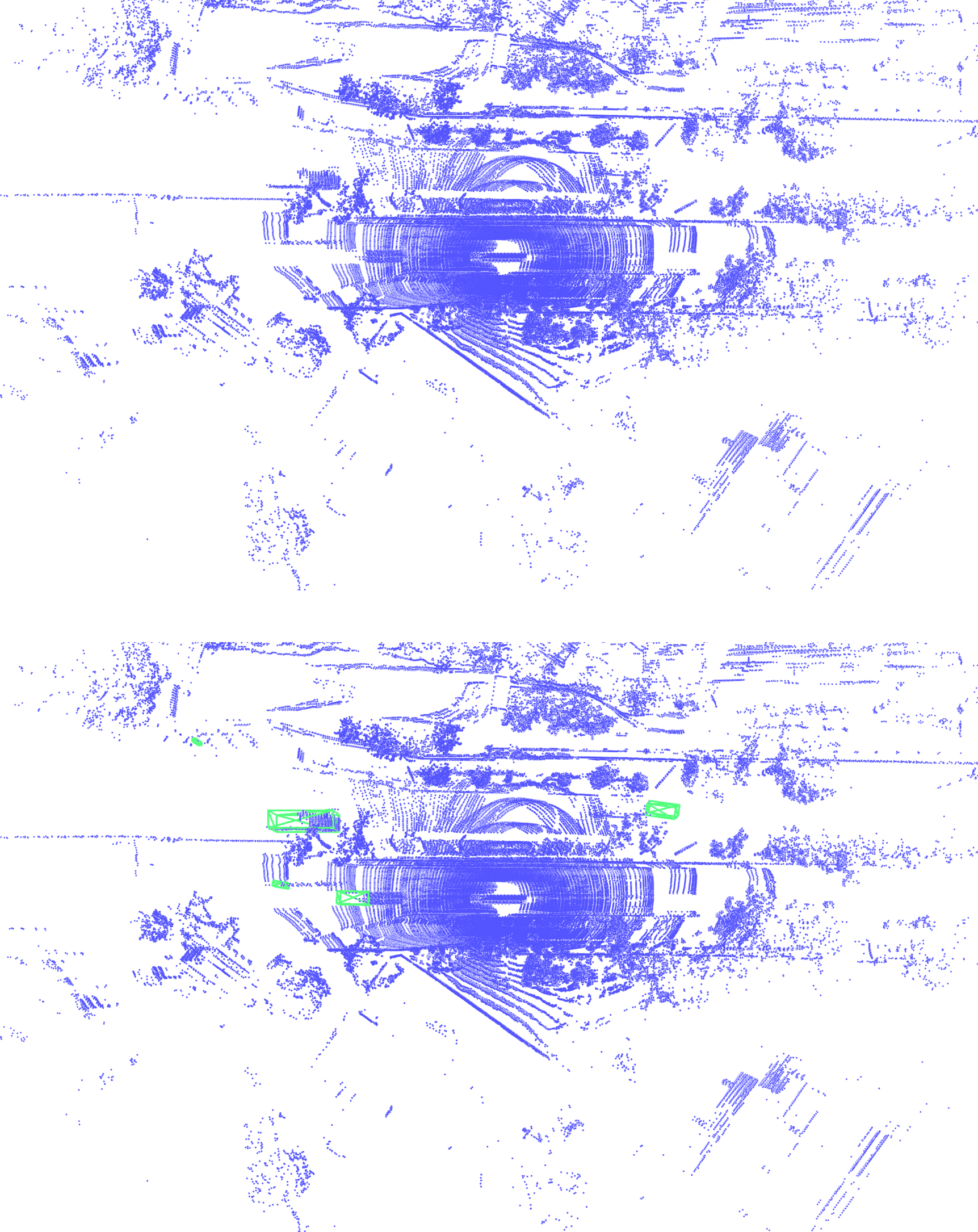}}
%\caption{fig2}
\end{minipage}
}%
\subfigure[Scene4]{
\begin{minipage}[t]{0.25\linewidth}
\centerline{\includegraphics[width=5cm]{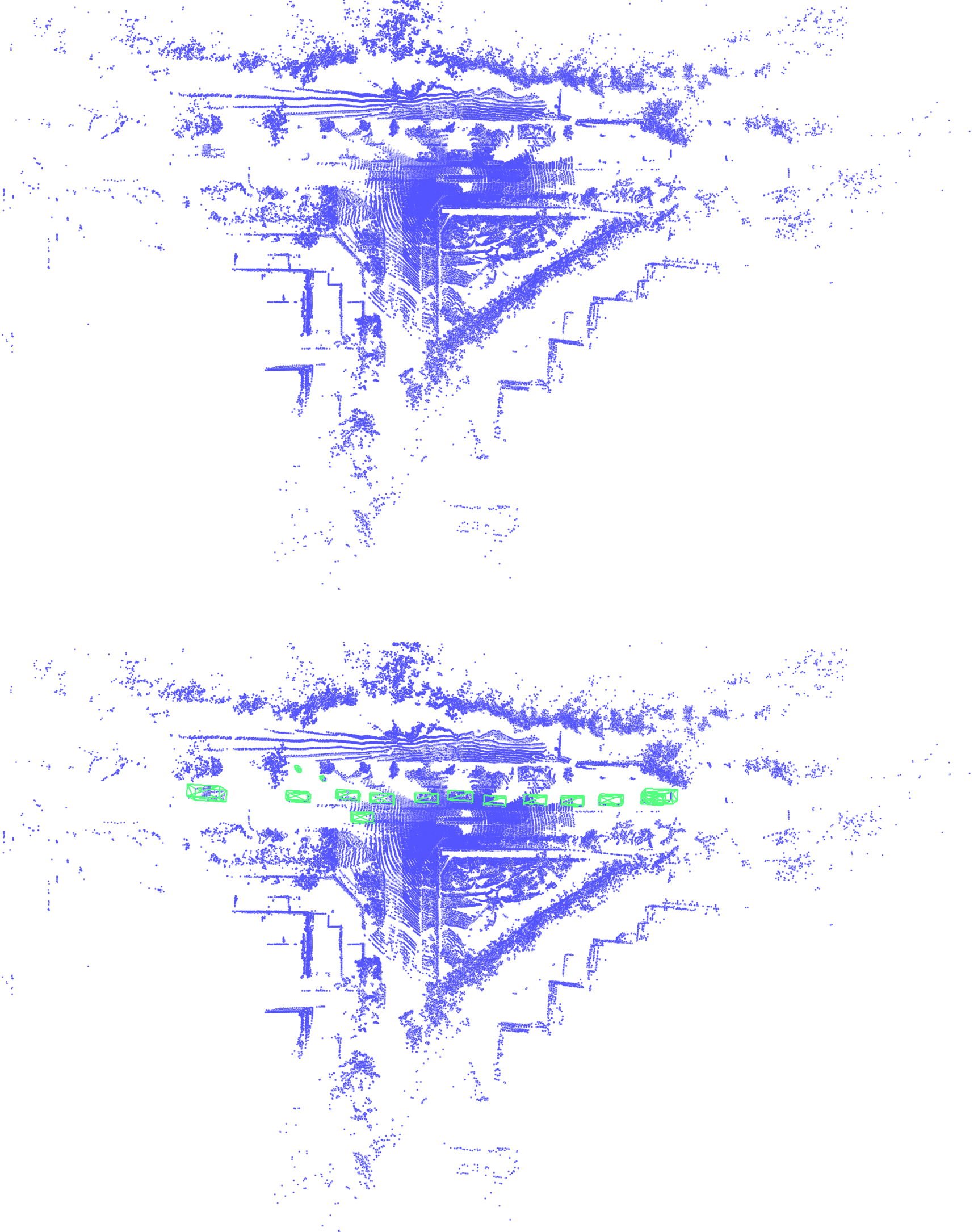}}
%\caption{fig2}
\end{minipage}
}%

\caption{3D Object detection visualization performed by GeqBevNet.}
\label{fig3}
\end{figure*}

\subsection{BEV-level Equivariant Feature Extraction}
In order to obtain the dense semantic information from the camera and the accurate geometric information from the LiDAR simultaneously, as shown in Fig.~\ref{fig1}, we use BEV to unify the information from both modalities into the same form. Two classical works \cite{b14}, \cite{b15} based on BEV fusion have inspired us to construct BEV fusion object detection network. In the camera BEV generation network, the backbone network uses DB-Swin-T \cite{b36} and the neck network uses FPN \cite{b32}. Futher, we encode the original surround view into BEV with LSS \cite{b24}, where the depth prediction module predicts the depth with high reliability due to the strong prior. Meanwhile, in the LiDAR BEV generation network, the original point cloud is first processed into regular voxels using Voxelnet \cite{b33}, compared with the faster Pointpillars \cite{b34}, Voxels can retain more spatial information of point clouds. Then, we use SECOND \cite{b46} as the backbone network for point cloud feature extraction and FPN \cite{b32} as the neck network. In addition, after fusing information from the camera and LiDAR on the BEV, channel feature augmentation is performed using the SE mechanism \cite{b50}, and the transfusion \cite{b47} is used as the detection head in order to efficiently detect the objects in the fused BEV.

After the image BEV and point cloud BEV are fused, we construct an embedded group equivariant network to extract the fused equivariant features. As shown in Fig.~\ref{fig2}. The above fusion process can be expressed as:
\begin{equation}
    F_{bevf}=f_{se}[f_{concat}(F_{bevi}(x),F_{bevp}(x))]  \label{eq5}
\end{equation}
where $F_{bevi}(x)$ denotes the image BEV features, $F_{bevp}(x)$ denotes the point cloud BEV features, $f_{concat}$ denotes the fusion BEV and $f_{se}$ denotes SE mechanism. Meanwhile, the constructed group equivariant network can be expressed as:
\begin{equation}\label{eq6}
\begin{aligned}
N_{geqbev}&=P(F_a[BN_{beveq}[GeqConV[F_{bevf}(x)]]])\\
&=P(F_{g}(x))\\
\end{aligned}
\end{equation}
where $GeqConV$ is consists of BEV-level lifting layer and BEV group convolution layer, $BN_{beveq} $ is the batch normalization that satisfies group $G$, $F_a$ is the activation function of the CNN, $P$ denotes the pooling layer we construct and $F_{g}(x)$ is the output of the activation function.

Contrary to the traditional CNN, group equivariant convolutional neural networks require to lift features from plane $\mathbb{Z}^{2}  $  to group $G$. To satisfy this definition, according to \eqref{eq3}, we construct a BEV-level lifting layer is called $BEVLift$ and express as follows:
\begin{equation}\label{eq7}
    \begin{aligned}
F_{bevlift}&=BEVLift(F_{bevf})\\
&= {[F_{bevf} \star  \psi](g) } \\
& =\sum_{n \in \mathbb{Z}^{2}} \sum_{i} F_{bevf i}(n) \psi_{i}\left(g^{-1} n\right)\\
\end{aligned}
\end{equation}
where $F_{bevf}$ is denotes the fused BEV feature map , $\psi$ is the group convolution kernel with both translation and rotation properties, $i$ is represents the number of channels and $n$ belongs to plane $\mathbb{Z} ^{2}$. Further, the BEV features initially on the plane $\mathbb{Z} ^{2}$ are lifted by one dimension after passing through the $BEVLift$. Therefore, the BEV features are lifted from $(B,C,H,W) $ to $(B,C,R,H,W)$, and to satisfy the definition of convolution on the group, according to \eqref{eq4}, we build the BEV group convolution as follows:

\begin{equation}\label{eq8}
\begin{aligned}
F_{geqconv}&=ConV_{beveq} (F_{bevlift})\\
&=[F_{bevlift}\star \psi](g)\\
& =\sum_{m \in G} \sum_{i} F_{bevlift i}(m) \psi_{i}\left(g^{-1} m\right)
\end{aligned}
\end{equation}
where $F_{bevlift}$ is the output of $BEVLift$, and $m$ belongs to group $g$, we use the cycle group $C4$ as the transform group with $R$ in the shape of BEV as 4. After the $BEVLift$ layer and the BEV group convolutional layer, we extract BEV-level rotational equivariant features from the BEV located in the plane $\mathbb{Z} ^{2}$.

\subsection{BEV-level Equivariant Feature Pooling}
Since the existing object detection head cannot directly detect objects and predict their orientation in the group, if we modify the existing object detection head to satisfy the rotational equivariant feature, it cannot satisfy the embeddedness requirement of the group equivariant network we designed. Therefore, to address this issue, we degenerate it from the group to the plane $\mathbb{Z}^{2} $. The features on the group have both translational and rotational equivariance, while the traditional CNN only has translation equivariant. When the shape of the features on the group corresponds to the shape of the features on the plane $\mathbb{Z}^{2}$, we call it degenerating from the group to the plane $\mathbb{Z}^{2}$. We build the $BEVEqPooling$ layer to eliminate the C4 dimension, which can be expressed as: 
\begin{equation}\label{eq9}
    \begin{aligned}
F_{geqbev}&=BEVEqPooling(F_{g}(x)) \\
\end{aligned}
\end{equation}

By building the $BEVEqPooling$ layer, we transfer the features from the group to the plane $\mathbb{Z}^{2}$, satisfying the requirements of embeddable group equivariant network.

%第四个表
\begin{table*}[htbp]

\caption{Comparison with other methods (mAP and NDS) On NuScenes Validation Set. }
\centering
\begin{center}

\begin{tabular}{|c|c|c|c|c|c|c|c|c|c|c|c|c|c|}
\hline
\textbf{}&\textbf{}&\multicolumn{10}{|c|}{\textbf{Per-class}}&\multicolumn{2}{|c|}{\textbf{Metric}} \\
\cline{3-14} 
\textbf{Method} & \textbf{\textit{Modality}}& \textbf{\textit{Car}}& \textbf{\textit{Trunk}}& \textbf{\textit{Bus}}& \textbf{\textit{Trailer}}& \textbf{\textit{C.V.}}& \textbf{\textit{Ped.}}& \textbf{\textit{Motor.}}& \textbf{\textit{Bicycle}}& \textbf{\textit{T.C.}}& \textbf{\textit{Barrier}}& \textbf{\textit{mAP$\uparrow $}}& \textbf{\textit{NDS$\uparrow $}} \\
\hline
PointPillars\cite{b34}&L&78.7&37.2&49.7&26.2&6.56&61.2&20.2&0.85&18.9&41.4&34.1&49.9\\
FreeAnchor\cite{b9}&L&81.2&39.3&48.0&30.9&10.2&74.4&43.5&18.0&41.4&52.7&44.0&55.0\\
CenterPoint\cite{b25}&L&83.9&50.2&62.0&32.7&10.5&77.3&45.4&16.4&50.5&60.1&48.9&59.6\\
HotSpotNet\cite{b49}&L&84.0&56.2&67.4&38.0&20.7&82.6&66.2&49.7&65.8&64.3&59.5&66.0\\
\hline
PointPainting\cite{b11}&L+C&77.9&35.8&36.2&37.3&15.8&73.3&41.5&24.1&62.4&60.2&46.4&58.1\\
3D CVF\cite{b13}&L+C&83.0&45.0&48.8&\textbf{49.6}&15.9&74.2&51.2&30.4&62.9&65.9&52.7&62.3\\
MSF3DDETR\cite{b37}&L+C&\textbf{86.0}&58.0&\textbf{71.0}&40.0&21.0&83.0&67.0&\textbf{53.0}&67.0&61.0&60.7&66.7\\
\hline
GeqBevNet(ours)&L+C&85.6&\textbf{58.2}&70.8&39.0&\textbf{24.8}&\textbf{85.3}&\textbf{67.2}&51.8&\textbf{69.6}&\textbf{70.0}&\textbf{62.2}&\textbf{66.8}\\
\hline
\multicolumn{14}{l}{Class names for class abbreviations: Construction Vehicle (C.V.), Pedestrian (Ped.), Traffic Cone (T.C.), Camera (C), LiDAR (L).}

\end{tabular}
\label{tab1}
\end{center}

\end{table*}

\section{Experiments}
To evaluate the performance of our proposed GeqBevNet, we conducted extensive experiments on the nuScenes dataset with different road scenes.
\subsection{Datasets}
Since BEV generation requires a complete vehicle surround view, KITTI \cite{b44} only provides the front view. Waymo \cite{b45} has $5$ cameras and lacks the camera information behind the vehicle, while nuScenes provides $6$ complete vehicles surrounding information cameras and becomes the most popular public dataset for BEV generation. In addition to the $6$ cameras, the nuScenes dataset is equipped with $32$-line LiDAR and $5$ millimeter-wave radars, and is equipped with an IMU.

Nuscenes dataset only selects key frames for labeling, and Boston and Singapore collected $40,000$ key frames. A total of $23$ classes of objects are labeled with 3D boxes, class information, and some important attributes. In terms of object detection, it supports 10 types of object detection, such as cars, trucks, pedestrians, fences, etc.

In object detection, the following types of evaluation metrics are available:  Average Precision (AP), Average Translation Error (ATE), Average Scale Error (ASE), Average Orientation Error (AOE), Average Velocity Error (AVE), and Average Attribute Error (AAE). In addition, the nuScenes dataset also proposes a nuScenes detection score (NDS), which is calculated using the six metrics mentioned above.

\subsection{Experimental Settings}
We will focus on the three metrics of mAP, mAOE and NDS, and demonstrate the effectiveness of our network by comparing these three metrics with other advanced networks. We set the input image to $1600\times 900$, and the point cloud range to $[-54.0, -54.0, -5.0, 54.0, 54.0, 3.0]$. In the training phase, in order to keep the iteration of the network stable, we added pre-trained weights, and then used the Tesla V100S to iterate the entire network for 6 epochs. The batch size of each epoch is set to 1, and the learning rate of the Adam optimizer is set to $0.001$. In addition, CBGS \cite{b51} is used as the sampling strategy in training the LiDAR stream, and the fade strategy is used. The latter means that Pointaugmenting \cite{b52} is used in the first 15 epochs, while this augmentation is deactivated in the last 5.

\subsection{Results} % Analysis of 
In this section, we show the performance achieved on the nuScenes validation set and compare it with advanced object detection methods. We provide visualizations of 3D object detection results performed by GeqBevNet in different scenarios in Fig.~\ref{fig3}. It shows that the GeqBevNet achieves promising performance in object orientation.

\subsubsection{Precision}
We compare the precision and NDS of our network with other advanced networks and place the results in Table \ref{tab1}. GeqBevNet achieved promising results without using other data augmentation methods. We achieved $62.2\%$ in mAP and $66.8\%$ in NDS. Through the above results, we fully demonstrate the effectiveness of the embedded group equivariant network we constructed.

\subsubsection{Orientation}
Since group equivariant networks can extract more rotational equivariant features, to verify the performance of GeqBevNet in object orientation prediction, we evaluate the network's AOE metrics for $9$ classes and compare them with advanced networks. The results are shown in Table \ref{tab2}, in which Our network decreases to 0.325 on mAOE. And from Table \ref{tab2} that the group equivariant network is more effective in predicting object orientation.

%第五个表
\begin{table}[htbp]
\caption{Comparison with other methods (AOE) On NuScenes Validation Set. }
\centering
\begin{center}
\begin{tabular}{|c|c|c|}
\hline
\textbf{}&\textbf{}&\multicolumn{1}{|c|}{\textbf{Metric}} \\
\cline{3-3}
\textbf{Method} & \textbf{\textit{Modality}}&  \textbf{\textit{mAOE$\downarrow $}} \\
\hline
DETR3D\cite{b16}&C&0.379\\
BEVFormer\cite{b29}&C&0.372\\
\hline
FreeAnchor\cite{b9}&L&0.530\\
PointPillars\cite{b34}&L&0.523\\
CenterPoint\cite{b25}&L&0.385\\
\hline
CenterFusion\cite{b4}&C+R&0.535\\
RCBEV4d\cite{b5}&C+R&0.445\\
\hline
BEVDepth\cite{b26}&L(S)+C&0.358\\

\hline
GeqBevNet(ours)&L+C&\textbf{0.325}\\
\hline
\multicolumn{3}{l}{ Radar(R),LiDAR Supervision(L(S)).}
\end{tabular}
\label{tab2}
\end{center}

\end{table}

%第六个表
\begin{table*}[htbp]
\caption{Results Of Group Equivariant Networks With Different Network Depths On NuScenes Validation Set.}
\centering
\begin{center}
\begin{tabular}{|c|c|c|c|c|c|c|c|c|c|c|}
\hline
\textbf{Network}&\multicolumn{9}{|c|}{\textbf{Per-class}}&\multicolumn{1}{|c|}{\textbf{Metric}} \\
\cline{2-11} 
\textbf{Depth} & \textbf{\textit{Car}}& \textbf{\textit{Trunk}}& \textbf{\textit{Bus}}& \textbf{\textit{Trailer}}& \textbf{\textit{C.V.}}& \textbf{\textit{Ped.}}& \textbf{\textit{Motor.}}& \textbf{\textit{Bicycle}}& \textbf{\textit{Barrier}}& \textbf{\textit{mAOE$\downarrow $}} \\
\hline
2&0.121&0.118&0.096&0.598&0.973&0.387&\textbf{0.247}&0.466&0.116& 0.347  \\
3&\textbf{0.119}&\textbf{0.105}&\textbf{0.071}&\textbf{0.537}&\textbf{0.914}&0.384&0.286&\textbf{0.394}&\textbf{0.112}& \textbf{0.325}  \\
4&0.120&0.130&0.133&0.733&0.984&\textbf{0.376}&0.299&0.459&0.130& 0.374  \\
\hline
% \multicolumn{4}{l}{$^{\mathrm{a}}$Sample of a Table footnote.}
\end{tabular}
\label{tab3}
\end{center}
\end{table*}

\begin{table*}[htbp]
\caption{Comparison Results Of Different Pooling Methods On Nuscenes Validation Set.}
\centering
\begin{center}
\begin{tabular}{|c|c|c|c|c|c|c|c|c|c|c|}
\hline
\textbf{Pooling}&\multicolumn{9}{|c|}{\textbf{Per-class}}&\multicolumn{1}{|c|}{\textbf{Metric}} \\
\cline{2-11} 
\textbf{Methods} & \textbf{\textit{Car}}& \textbf{\textit{Trunk}}& \textbf{\textit{Bus}}& \textbf{\textit{Trailer}}& \textbf{\textit{C.V.}}& \textbf{\textit{Ped.}}& \textbf{\textit{Motor.}}& \textbf{\textit{Bicycle}}&  \textbf{\textit{Barrier}}& \textbf{\textit{mAOE$\downarrow $}} \\
\hline
Max Pooling&0.247&0.309&0.411&0.902&1.028&0.775&0.642&0.984&0.336&0.626\\
Average Pooling&\textbf{0.112}&0.110&0.078&0.593&\textbf{0.898}&\textbf{0.370}&\textbf{0.266}&0.490&0.115&0.337\\
BevEq Pooling&0.119&\textbf{0.105}&\textbf{0.071}&\textbf{0.537}&0.914&0.384&0.286&\textbf{0.394}&\textbf{0.112}& \textbf{0.325}  \\

\hline
% \multicolumn{4}{l}{$^{\mathrm{a}}$Sample of a Table footnote.}
\end{tabular}
\label{tab4}
\end{center}
\end{table*}

\subsection{Ablation}
To demonstrate the effect of group equivariant network depth on the overall network,
% group equivariant network depth on the effectiveness of BEV fused object detection networks, 
We conduct ablation experiments on the depth of group equivariant networks to obtain an optimal network depth.

From Section \ref{sec_3}, we know that the group equivariant network in GeqBevNet consists of $BEVLift$ layer, BEV group convolutional layer, and $BEVEqPooling$ layer. Among them, the $BEVLift$ layer and $BEVEqPooling$ layer do not require to increase or decrease the number of layers due to their particular functions, and we pay more attention to the BEV group convolutional layer, especially on its number of layers. We will choose the optimal number of layers $N$ to obtain the best orientation prediction performance based on the lowest mAOE of the network.

We set up the group equivariant network with different depths in GeqBevNet and conducted experiments to obtain the results shown in Table \ref{tab3}. We can observe that when $N=3$, the network achieves the best results on mAOE. After $N=3$, mAOE gradually increases. According to the ablation experiments, the depth of the group equivariant network does not mean that extracting deeper rotational equivariant features can improve the network's orientation prediction performance.  In addition, there may be insufficient feature extraction for shallow rotational equivariant features, which still cannot improve the average orientation error of the object.

Further, in order to prove the effectiveness of our constructed $BEVEqPooling$ layer, we conducted ablation experiments with average pooling and max pooling, and the obtained results are shown in Table \ref{tab4}. We can observe that our built $BEVEqPooling$ layer is effective in decreasing the AOE of the network compared to the average pooling and max pooling.

\section{CONCLUSION}
In this paper, we propose GeqBevNet, in which a group equivariant network is embedded into the object detection network to address the perturbations of 3D objects under different driving conditions. The group equivariant network enables the extraction of BEV-level rotational equivariant features to obtain better performance in object detection tasks. In addition, the constructed group equivariant network can be easily embedded into other end-to-end networks to increase their orientation prediction performance. 

Since the rotation of the actual road scene may be more sophisticated, there are many rotation factors that we may have not considered. Currently, we only consider the issue of large-scale scene rotation. In the future, we will continue to explore the existing problems and extend the group equivariant network to more complex and applicable specific discrete groups, focusing on the issues caused by road scene rotation.

\section*{Acknowledgment}
This research is supported by the National Natural Science Foundation of China (No.42130112) and KartoBit Research Network(No.KRN2201CA). Partially supported by ‘Fujian Science \& Technology Innovation Laboratory for Optoelectronic Information of China’ (No.2021ZZ120). The authors would like to thank Xihao Wang and Jiaming Lei for their helpful discussion.

%第七个文献

\vspace{12pt}

\end{document}